# A Semantic Similarity Measure for Expressive Description Logics

Claudia d'Amato, Nicola Fanizzi, Floriana Esposito

Dipartimento di Informatica, Università degli Studi di Bari
Campus Universitario, Via Orabona 4, 70125 Bari, Italy
{claudia.damato, fanizzi, esposito}@di.uniba.it

**Abstract.** A totally semantic measure is presented which is able to calculate a similarity value between concept descriptions and also between concept description and individual or between individuals expressed in an expressive description logic. It is applicable on symbolic descriptions although it uses a numeric approach for the calculus. Considering that Description Logics stand as the theoretic framework for the ontological knowledge representation and reasoning, the proposed measure can be effectively used for agglomerative and divisional clustering task applied to the semantic web domain.

## 1 Introduction

Ontological knowledge plays a key role for interoperability in the Semantic Web perspective. Nowadays, standard ontology markup languages are supported by well-founded semantics of Description Logics (DLs) together with a series of available automated reasoning services [1]. However, several tasks in an ontology life-cycle [2], such as their construction and/or integration, are still almost entirely delegated to knowledge engineers.

In the Semantic Web perspective, the construction of the knowledge bases should be supported by automated inductive inference services. The induction of structural knowledge like the T-box taxonomies is not new in machine learning, especially in the context of *concept formation* [3] where clusters of similar objects are aggregated in hierarchies according to heuristic criteria or similarity measures. Almost all of these methods apply to zero-order representations while, as mentioned above, ontologies are expressed through fragments of first-order logic. Yet, the problem of the induction of structural knowledge turns out to be hard in first-order logic or equivalent representations [4].

In *Inductive Logic Programming* (ILP) attempts have been made to extend relational learning techniques towards hybrid representations based on both clausal and description logics [5, 6, 7]. In order to cope with the problem complexity, these methods are based on a heuristic search and generally implement bottom-up algorithms that tend to induce overly specific concept definitions which may suffer for poor predictive capabilities.

So far, the automated induction of knowledge bases expressed in DLs representations has not been investigated in depth. Classic approaches to learning

DL concept definitions generally adopt heuristic search strategies to cope with the inherent complexity of the problem and generally implement bottom-up algorithms (e.g. [8]). Other approaches propose a top-down search for correct concept definitions [9]. These methods are not completely operational: since refinement operators compute short moves in a vast space of candidate definitions, they become useless when disjoined from proper heuristics based on the available assertions. A more knowledge-intensive method is to be preferred.

In this perspective, we introduce a novel similarity measure between concept descriptions based on semantics, which is suitable for expressive DLs like $\mathcal{ALC}$ [10, 1]. Since a merely syntactic approach has proven too weak to enforce standard inferences (namely subsumption), when expressive DLs are taken into account a different approach (based on semantics) is necessary. Also a similarity measure, then, should be founded on the underlying semantics, rather than on the syntactic structure of concept descriptions. Besides measuring the similarity of two concept descriptions, we propose a manner based on notion of *most specific concept* of an individual [1] for employing the same measure for the individual-concept and individual-individual similarity cases.

Such a measure can be the basis for adapting an existing clustering method to this representation (or devising a new one) operating in a top-down (*partitional*) or bottom-up (*agglomerative*) fashion. Moreover, the similarity measure can be also employed for *Information Retrieval* or *Information Integration* purposes applied to DL knowledge bases and also for *Case-based Reasoning* systems (see the next section).

As discussed in the following, the method can effectively compute the similarity measure with a complexity which depends on the complexity of standard inferences as a baseline. The applicability of this has been tested on examples which have been artificially generated from OWL[1] ontologies. Some of these test examples are reported in this paper.

The remainder of the paper is organized as follows. The next section reviews related work on similarity measures. In Sect. 3 the representation language is presented. The similarity measure is illustrated in Sect. 4 and is discussed in Sect. 5. Possible developments of the method are examined in Sect. 6.

## 2 Related Work

Similarity measure play an important role in information retrieval and information integration. Recent investigations in these fields have emphasized the use of ontologies and semantic similarity functions as a mechanism for comparing concepts and/or concept instances that can be retrieved or integrated across heterogeneous repositories [11, 12, 13, 14].

Semantic similarity is typically determined as a function of the *path distance* between terms in the hierarchical structure underlying the ontology [15, 16, 17]. Other methods to assess semantic similarity within a single ontology are *feature*

[1] `http://www.w3.org/TR/owl-ref`

*matching* [19] and *information content* [13, 20]. The former approach uses both common and discriminant features among concepts and/or concept instances in order to compute the semantic similarity. The latter methods are founded on *Information Theory*. They define a similarity measure between two concepts within a concept hierarchy in terms of the amount of information conveyed by the immediate super-concept that subsumes two concepts being compared. This is a measure of the variation of information crossing from a description level to a more general one.

A recent work [21] presents a number of measures for comparing concepts located in different and possibly heterogeneous ontologies. The following requirements are made for this measure:

- the formal representation supports inferences such as *subsumption*;
- local concepts in different ontologies inherit their definitional structure from concepts in a shared ontology.

This study assumes that the intersection of sets of concept instances is an indication of the correspondence between these concepts. Three main types of measures for comparing concept descriptions are discussed in this work:

1. *filter* measures based on a path-distance
2. *matching* measures based on graph matching establish one-to-one correspondence between elements of the concept descriptions, and
3. *probabilistic* measures that give the correspondence in terms of the joint distribution of concepts.

Other similarity measures have been developed to compute similarity values among classes belonging to different ontologies. These measures are able to take into account the difference in the levels of explicitness and formalization of the different ontology specifications. Particularly, in [22] a similarity function determines similar entity classes by using a matching process making use of synonym sets, semantic neighborhood, and discriminating features that are classified into parts, functions, and attributes.

Another approach [23], aimed at finding commonalities among concepts or among assertions, employs the *Least Specific Concept* operator (LCS [1]) that computes the most specific generalization of the input concepts (with respect to subsumption, see the next section for a formal definition). This approach is generally intended for information retrieval purposes. Considered a knowledge base and a query concept, a filter mechanism selects another concept from the knowledge base that is relevant for the query concept. Then the LCS of the two concepts is computed and finally all concepts subsumed by the LCS are returned.

Most of the cited works adopt a semantic approach in conjunction with the structure of the considered concept descriptions. Thus, they are liable to the phenomenon of the rapid growth of the description granularity. Besides the syntactic structure of concept descriptions becomes much less important when richer DL representations are adopted due to the expressive operators that can be employed. For these reasons, we have decided to focus our attention to a measure which is totally based on semantics.

**Table 1.** $\mathcal{ALC}$ constructors and their meaning.

| Name | Syntax | Semantics |
|---|---|---|
| top concept | $\top$ | $\Delta^{\mathcal{I}}$ |
| bottom concept | $\bot$ | $\emptyset$ |
| concept | $C$ | $C^{\mathcal{I}} \subseteq \Delta^{\mathcal{I}}$ |
| concept negation | $\neg C$ | $\Delta^{\mathcal{I}} \setminus C^{\mathcal{I}}$ |
| concept conjunction | $C_1 \sqcap C_2$ | $C_1^{\mathcal{I}} \cap C_2^{\mathcal{I}}$ |
| concept disjunction | $C_1 \sqcup C_2$ | $C_1^{\mathcal{I}} \cup C_2^{\mathcal{I}}$ |
| existential restriction | $\exists R.C$ | $\{x \in \Delta^{\mathcal{I}} \mid \exists y \in \Delta^{\mathcal{I}}((x,y) \in R^{\mathcal{I}} \wedge y \in C^{\mathcal{I}})\}$ |
| universal restriction | $\forall R.C$ | $\{x \in \Delta^{\mathcal{I}} \mid \forall y \in \Delta^{\mathcal{I}}((x,y) \in R^{\mathcal{I}} \rightarrow y \in C^{\mathcal{I}})\}$ |

## 3 The Reference Representation Language

In relational learning, several solutions have been proposed for the adoption of an expressive fragment of first-order logic endowed with efficient inference procedures. Alternatively, the data model of a knowledge base can be expressed by means of DL concept languages which are empowered with precise semantics and effective inference services [1]. Besides, most of the ontology markup languages for the Semantic Web (e.g., OWL) are founded in Description Logics: representation languages borrow and implement the typical constructors of the DL languages.

Although it can be assumed that annotations and conceptual models are maintained and transported using the XML-based languages mentioned above, the syntax of the representation adopted here is taken from the standard constructors proposed in the DL literature [1]. These DL representations turn out to be both sufficiently expressive and efficient from an inferential viewpoint.

In this section we recall syntax and semantics for the reference representation $\mathcal{ALC}$ [10] which is adopted in the rest paper for it turns out to be sufficiently expressive to support most of the principal constructors of an ontology markup language for the Semantic Web.

In a DL language, primitive *concepts*, denoted with names taken from $N_C = \{C, D, \ldots\}$, are interpreted as subsets of a certain domain of objects (resources) or equivantly as unary relation on such domain and primitive *roles*, denoted with names taken from $N_R = \{R, S, \ldots\}$, are interpreted as binary relations on such a domain (properties). Complex concept descriptions can be built using primitive concepts and roles by means of the constructors in Table 1. Their semantics is defined by an *interpretation* $\mathcal{I} = (\Delta^{\mathcal{I}}, \cdot^{\mathcal{I}})$, where $\Delta^{\mathcal{I}}$ is the *domain* of the interpretation and the functor $\cdot^{\mathcal{I}}$ stands for the *interpretation function*, mapping the intension of concepts and roles to their extension.

A *knowledge base* $\mathcal{K} = \langle \mathcal{T}, \mathcal{A} \rangle$ contains two components: A *T-box* $\mathcal{T}$ and an *A-box* $\mathcal{A}$. $\mathcal{T}$ is a set of concept definitions $C \equiv D$, meaning $C^{\mathcal{I}} = D^{\mathcal{I}}$, where $C$ is the concept name and $D$ is a description given in terms of the language constructors. $\mathcal{A}$ contains extensional assertions on concepts and roles, e.g. $C(a)$ and $R(a, b)$, meaning, respectively, that $a^{\mathcal{I}} \in C^{\mathcal{I}}$ and $(a^{\mathcal{I}}, b^{\mathcal{I}}) \in R^{\mathcal{I}}$; $C(a)$ and $R(a, b)$ are said respectively instance of the concept $C$ and instance of the role $R$, more generally it is said (without loss of generality) that the individual $a$ is instance of the concept $C$ and the same for the role. A notion of *subsumption* between concepts is given in terms of the interpretations:

**Definition 3.1 (subsumption).** *Given two concept descriptions $C$ and $D$, $C$* subsumes *$D$, denoted by $C \sqsupseteq D$, iff for every interpretation $\mathcal{I}$ it holds that $C^{\mathcal{I}} \supseteq D^{\mathcal{I}}$.*

Axioms based on subsumption ($C \sqsupseteq D$) are generally also allowed in the T-boxes as partial definitions. Indeed, $C \equiv D$ amounts to $C \sqsupseteq D$ and $D \sqsupseteq C$.

*Example 3.1.* An instance of concept definition in the proposed language is:

$$\mathsf{Father} \equiv \mathsf{Male} \sqcap \exists \mathsf{hasChild}.\mathsf{Person}$$

which corresponds to the sentence: *"a father is a male (person) that has some persons as his children"*.

The following are instances of simple assertions:

$$\mathsf{Male}(\mathsf{Leonardo}),\ \mathsf{Male}(\mathsf{Vito}),\ \mathsf{hasChild}(\mathsf{Leonardo}, \mathsf{Vito}).$$

Supposing that $\mathsf{Male} \sqsubseteq \mathsf{Person}$ is known (in the T-Box), one can deduce that: $\mathsf{Person}(\mathsf{Leonardo})$, $\mathsf{Person}(\mathsf{Vito})$ and then $\mathsf{Father}(\mathsf{Leonardo})$.

Given these primitive concepts and roles, it is possible to define many other related concepts:

$$\mathsf{Parent} \equiv \mathsf{Person} \sqcap \exists \mathsf{hasChild}.\mathsf{Person}$$

and

$$\mathsf{FatherWithoutSons} \equiv \mathsf{Male} \sqcap \exists \mathsf{hasChild}.\mathsf{Person} \sqcap \forall \mathsf{hasChild}.(\neg \mathsf{Male})$$

It is easy to see that the following relationships hold: $\mathsf{Parent} \sqsupseteq \mathsf{Father}$ and $\mathsf{Father} \sqsupseteq \mathsf{FatherWithoutSons}$. □

Especially for rich DL languages such as $\mathcal{ALC}$, many semantically equivalent (yet syntactically different) descriptions can be given for the same concept, which is the reason for preferring employing semantic approaches to reasoning over structural ones. Nevertheless equivalent concepts can be reduced to a normal form by means of rewriting rules that preserve their equivalence, such as: $\forall R.C_1 \sqcap \forall R.C_2 \equiv \forall R.(C_1 \sqcap C_2)$ (see [1] for issues related to normalization and simplification).

One of the most important inference services from the inductive learning viewpoint is *instance checking*, that is deciding whether an individual is an instance of a concept (w.r.t. an A-Box). Related to this problem, it is often necessary to solve the *realization problem* that requires to compute, given an A-Box and an individual the concepts which the individual belongs to:

**Definition 3.2 (Most Specific Concept).** *Given an A-Box $\mathcal{A}$ and an individual $a$, the most specific concept of $a$ w.r.t. $\mathcal{A}$ is the concept $C$, denoted $\mathsf{MSC}_{\mathcal{A}}(a)$, such that $\mathcal{A} \models C(a)$ and $\forall D$ such that $\mathcal{A} \models D(a)$, it holds: $C \sqsubseteq D$.*

where $\models$ stands for the standard semantic deduction [24].

In the general case of a cyclic A-Box expressed in a an expressive DL endowed with existential or numeric restriction the $\mathsf{MSC}$ cannot be expressed as a finite concept description [1], thus it can only be approximated.

Since the existence of the $\mathsf{MSC}$ for an individual w.r.t. an A-Box is not guaranteed or it is difficult to compute, generally an approximation of the $\mathsf{MSC}$ is considered up to a certain depth $k$. The maximum depth $k$ has been shown to correspond to the depth of the considered A-Box, as defined in [23].

Henceforth we will indicate generically an approximation to the maximum depth with $\mathsf{MSC}^*$.

## 4 The Similarity Measure

In this section we present a similarity measure which is able to assess the similarity between instances or between instance and concept or even between concepts expressed in Description Logic and in particular in the $\mathcal{ALC}$ logic. We call such elements generically objects. Then we will formalize the measure for the various object types. The presented measure employs the basic set theory. It is mainly founded on the commonality among objects. Particularly, the base criterion for this measure is: the similarity value between objects is not only the result of the common features, but also the result of the different characteristics too. This criterion is in agreement with an information-theoretic definition of similarity [25].

### 4.1 A Similarity Measure between Concepts

Let $C$ and $D$ two concepts description in a T-Box, expressed in the $\mathcal{ALC}$ logic. Now recall that through the instance checking service it is possible to determine the set of all individuals of a given A-Box that are instances of a certain concept. Let $C^{\mathcal{I}}$ and $D^{\mathcal{I}}$ be, respectively, the extensions of the concepts $C$ and $D$ respectively. By Def. 3.1, $D$ subsumes $C$ (written $C \sqsubseteq D$) if the set of individuals that are instances of $C$ is contained in the set of individuals that are instances of $D$, in other words if $C^{\mathcal{I}} \subseteq D^{\mathcal{I}}$. Here we refer to the canonical interpretation of the A-Box and the *unique names assumption* (UNA) is made: constants in the A-Box are interpreted as themselves and different names for individuals stand for different domain objects (*canonical interpretation*).

Subsumption is a semantic relationship which induces an order over the space of concept descriptions: if $C \sqsubseteq D$ then $D$ is more general that $C$ or equivalently then $C$ is more specific then $D$. Based on subsumption and set theory, we define a semantic similarity measure:

**Definition 4.1 (Semantic Similarity Measure).** *Let $\mathcal{L}$ be the set of all concepts in $\mathcal{ALC}$ and let $\mathcal{A}$ be an A-Box with canonical interpretation $\mathcal{I}$. The* Semantic Similarity Measure *$s$ is a function*

$$s : \mathcal{L} \times \mathcal{L} \mapsto [0, 1]$$

*defined as follows:*

$$s(C, D) = \frac{|I^{\mathcal{I}}|}{|C^{\mathcal{I}}| + |D^{\mathcal{I}}| - |I^{\mathcal{I}}|} \cdot \max(|I^{\mathcal{I}}|/|C^{\mathcal{I}}|, |I^{\mathcal{I}}|/|D^{\mathcal{I}}|)$$

*where $I = C \sqcap D$ and $(\cdot)^{\mathcal{I}}$ computes the concept extension wrt the interpretation $\mathcal{I}$.*

The measure can be briefly justified as follows.
In case of semantic equivalence of the input concepts ($C \sqsubseteq D$ and $D \sqsubseteq C$), the maximum value of the similarity is assigned.

In case of disjunction, the minimum value of similarity is assigned because the two concepts are totally different: their extensions do not overlap. Indeed, they are semantically unrelated with respect to the generalization order: their intersection amounts to the bottom concept.

In case of overlapping concepts, a value in the range $]0, 1[$ is computed. It expresses the similarity between the two concepts (represented by the factor $|I^{\mathcal{I}}|/(|C^{\mathcal{I}}| + |D^{\mathcal{I}}| - |I^{\mathcal{I}}|)$) reduced by a quantity $(\max(|I^{\mathcal{I}}|/|C^{\mathcal{I}}|, |I^{\mathcal{I}}|/|D^{\mathcal{I}}|))$ which represents the major incidence of the intersection with respect to either concept. This means considering similarity not as an absolute value but as weighted with respect to a degree of non-similarity. Indeed, the higher such factor is the more one of the concepts is likely to be subsumed by the other. This is in accordance to the strong semantic relation between the concepts ensured by subsumption.

*Example 4.1.* Let be consider the knowledge base with the *T-Box* and *A-Box* reported below.

Primitive Concepts: $N_C = \{\mathsf{Female}, \mathsf{Male}, \mathsf{Human}\}$.

Primitive Roles: $N_R = \{\mathsf{HasChild}, \mathsf{HasParent}, \mathsf{HasGrandParent}, \mathsf{HasUncle}\}$.

T-Box: $\mathcal{T} = \{$
$\mathsf{Woman} \equiv \mathsf{Human} \sqcap \mathsf{Female}$
$\mathsf{Man} \equiv \mathsf{Human} \sqcap \mathsf{Male}$
$\mathsf{Parent} \equiv \mathsf{Human} \sqcap \exists \mathsf{HasChild}.\mathsf{Human}$

Mother ≡ Woman ⊓ Parent ∃HasChild.Human
Father ≡ Man ⊓ Parent
Child ≡ Human ⊓ ∃HasParent.Parent
Grandparent ≡ Parent ⊓ ∃HasChild.( ∃ HasChild.Human)
Sibling ≡ Child ⊓ ∃HasParent.( ∃ HasChild ≥ 2)
Niece ≡ Human ⊓ ∃HasGrandParent.Parent ⊔ ∃HasUncle.Uncle
Cousin ≡ Niece ⊓ ∃HasUncle.(∃ HasChild.Human)
}.

A-Box: $\mathcal{A}$ = {Woman(Claudia), Woman(Tiziana),
Father(Leonardo), Father(Antonio), Father(AntonioB),
Mother(Maria), Mother(Giovanna),
Child(Valentina),
Sibling(Martina), Sibling(Vito),
HasParent(Claudia,Giovanna), HasParent(Leonardo,AntonioB),
HasParent(Martina,Maria), HasParent(Giovanna,Antonio),
HasParent(Vito,AntonioB), HasParent(Tiziana,Giovanna),
HasParent(Tiziana,Leonardo), HasParent(Valentina,Maria),
HasParent(Maria,Antonio),
HasSibling(Leonardo,Vito), HasSibling(Martina,Valentina),
HasSibling(Giovanna,Maria), HasSibling(Vito,Leonardo),
HasSibling(Tiziana,Claudia), HasSibling(Valentina,Martina),
HasChild(Leonardo,Tiziana), HasChild(Antonio,Giovanna),
HasChild(Antonio,Maria), HasChild(Giovanna,Tiziana),
HasChild(Giovanna,Claudia), HasChild(AntonioB,Vito),
HasChild(AntonioB,Leonardo), HasChild(Maria,Valentina),
HasUncle(Martina,Giovanna), HasUncle(Valentina,Giovanna) }

Considered this knowledge base, it is possible to compute the similarity value between concepts as shown:

$$s(\mathsf{Grandparent}, \mathsf{Father}) = \frac{|\mathsf{Grandparent} \sqcap \mathsf{Father}|}{|\mathsf{Granparent}| + |\mathsf{Father}| - |\mathsf{Grandparent} \sqcap \mathsf{Father}|} \cdot$$
$$\cdot\ max\left(\frac{|\mathsf{Grandparent} \sqcap \mathsf{Father}|}{|\mathsf{Grandparent}|}, \frac{|\mathsf{Grandparent} \sqcap \mathsf{Father}|}{|\mathsf{Father}|}\right) = \frac{2}{2+3-2} \cdot max\left(\frac{2}{2}, \frac{2}{3}\right) = 0.67$$

In the same way it is possible to compute the similarity value among all concepts the defined above. □

### 4.2 Derived Similarity Measures Involving Individuals

Let us recall that, for every individual in the A-Box, it is possible to calculate the Most Specific Concept MSC (see Def. 3.2) or at least its approximation MSC*. In some cases they are equivalent concepts.

Let $a$ and $b$ two individuals in a given A-Box. We can calculate $A^* = \mathsf{MSC}^*(a)$ and $B^* = \mathsf{MSC}^*(b)$. Now the semantic similarity measure $s$ can be applied to these concept descriptions, thus yielding the similarity value of two instances:

$$\forall a, b: \ s(a, b) = s(A^*, B^*) = s(\mathsf{MSC}^*(a), \mathsf{MSC}^*(b))$$

Analogously, the similarity value between a concept description $C$ and an individual $a$ can be computed by determining the MSC approximation of the individual and then applying the similarity measure:

$$\forall a: \ s(a, C) = s(\mathsf{MSC}^*(a), C)$$

*Example 4.2.* Considering the knowledge base of the previous example, it is possible to show how to determine the similarity value between individuals. Let Claudia and Tiziana be such individuals. First of all, using the MSC* operator, we have:

MSC*(Claudia) = Woman ⊓ Sibling ⊓ ∃ HasParent(Mother ⊓ Sibling ⊓ ∃HasSibling(C1) ⊓ ∃HasParent(C2) ⊓ ∃HasChild(C3))

C1 ≡ Mother ⊓ Sibling ⊓ ∃HasParent(Father ⊓ Parent) ⊓ ∃HasChild(Cousin ⊓ ∃HasSibling(Cousin ⊓ Sibling ⊓ ∃HasSibling.⊤))

C2 ≡ Father ⊓ ∃HasChild(Mother ⊓ Sibling)

C3 ≡ Woman ⊓ Sibling ⊓ ∃HasSibling.⊤ ⊓ ∃HasParent(C4)

C4 ≡ Father ⊓ Sibling ⊓ ∃HasSibling(Uncle ⊓ Sibling ⊓ ∃HasParent(Father ⊓ Grandparent)) ⊓ ∃HasParent(Father ⊓ Grandparent ⊓ ∃HasChild(Uncle ⊓ Sibling))

And for the individual Tiziana:

MSC*(Tiziana) = Woman ⊓ Sibling ⊓ ∃HasSibling(Woman ⊓ Sibling ⊓ ∃HasParent(C5)) ⊓ ∃HasParent(Father ⊓ Sibling ⊓ ∃HasSibling(C6) ⊓ ∃HasParent(C7)) ⊓ ∃HasParent(Uncle ⊓ Mother ⊓ Sibling ⊓ ∃HasChild(Woman ⊓ Sibling))

C5 ≡ Mother ⊓ Sibling ⊓ ∃HasSibling(C8) ⊓ ∃HasParent(Father ⊓ ∃HasChild(Mother ⊓ Sibling))

C8 ≡ Mother ⊓ Sibling ⊓ ∃HasParent(Father ⊓ Grandparent) ⊓ ∃HasChild(Cousin ⊓ ∃HasSibling(Cousin ⊓ Sibling ⊓ ∃HasSibling.⊤))

C6 ≡ Uncle ⊓ Sibling ⊓ ∃HasParent(Father ⊓ Grandparent)

C7 ≡ Father ⊓ Granparent ⊓ ∃HasChild(Uncle ⊓ Sibling)

Note that it holds that $\mathsf{MSC}^*(\mathsf{Tiziana}) \not\sqsubseteq \mathsf{MSC}^*(\mathsf{Claudia})$ and $\mathsf{MSC}^*(\mathsf{Claudia}) \not\sqsubseteq \mathsf{MSC}^*(\mathsf{Tiziana})$. Now, since $\mathsf{MSC}^*(\mathsf{Tiziana}) = \{\mathsf{Tiziana}\}$ and $\mathsf{MSC}^*(\mathsf{Claudia}) =$

{Claudia, Tiziana}, the similarity value between these individuals is:

$$s(\mathsf{Claudia}, \mathsf{Tiziana}) = \frac{1}{1+2-1} \cdot max\left(\frac{1}{2}, \frac{1}{1}\right) = 0.5$$

In the same way it could be calculate the similarity value between concept and individual. □

## 5 Discussion

First of all it is important to note that, differently from previously proposed similarity measures (see Sect. 2), this measure is totally semantic. Indeed, it uses only semantic inferences like instance checking (to solve the *retrieval problem* [1], that amounts to computing the extension of a concept given an A-Box); it does not make use of the syntactic structure of the concept description, thus it is independent from the granularity level of descriptions. This fact reflects the intrinsic complexity of expressive DL languages like $\mathcal{ALC}$ for which a structural approach to reasoning is simply ineffective (subsumption is computed using a tableaux rather than a structural algorithm). Therefore, the definition of $s$ employs set theory and semantic services, so it make use of numeric approach despite its application on a symbolic DL representation.

Our similarity measure has been applied on knowledge base written using the $\mathcal{ALC}$ logic. However, for the reasons mentioned above, it is important to note that $s$ is applicable to any DL endowed with the basic reasoning services required by its definitions, namely: instance checking and MSC (approximation).

Similarity measures turn out to be useful in several applications and for many tasks such as commonality-based information retrieval in the context of a terminological knowledge representation system (which is a relatively new application context [23]), realization of semantic search engine, classification, case-based reasoning, clustering.

This last task is our goal. In particular, having defined a measure that is applicable both between concepts and between individuals and between concept and individuals, it is suitable for agglomerative clustering and for divisional clustering too. However we have noticed that $s$ measure is suitable for measuring similarity between concepts but it presents some problem in case of individuals. This is due to the use of the individual's MSC (approximation) which often turn out to be so specific that its extension likely includes only the considered individual; this phenomenon consequently provokes a totally dissimilarity value even if the individuals semantically express similar underlying concepts. So now we are investigating ways to overcome this limitation.

Below we prove that the function $s$ presented is really a similarity measure and discuss the complexity issues related to its computation.

### 5.1 Properties of the Similarity Measure

In this section we prove that $s$ function actually is a similarity measure (or *similarity function* [26]), according to the formal definition:

**Definition 5.1 (Similarity Measure).** *Let* E *be a set of elements among which a similarity measure has to be defined. A* similarity measure $E$ *is a real-valued function* $d$ *defined on the set* $E \times E$ *that fulfills the following properties:*

1. $f(a, b) \geq 0 \quad \forall a, b \in E \quad$ (positive definiteness)
2. $f(a, b) = f(b, a) \quad \forall a, b \in E \quad$ (symmetry)
3. $\forall a, b \in E : f(a, b) \leq f(a, a)$

¿From the definition given in the previous section it is straightforward to prove that $s$ satisfies the first property because $s$ has value in the real interval $[0, 1]$. Then, as previously said, $s$ assigns the maximum value when the concepts subsume each other; this last is the condition of equality of concept, so the third property is satisfied too. The property of symmetry is also trivially verified. Indeed set intersection, sum, product and maximum are commutative. It is straightforward to note that given two concepts $C$ and $D$, it holds that:

$$s(C, D) = \frac{|I^{\mathcal{I}}|}{|C^{\mathcal{I}}|+|D^{\mathcal{I}}|-|I^{\mathcal{I}}|} \cdot \max\left(\frac{|I^{\mathcal{I}}|}{|C^{\mathcal{I}}|}, \frac{|I^{\mathcal{I}}|}{|D^{\mathcal{I}}|}\right) = \frac{|I^{\mathcal{I}}|}{|D^{\mathcal{I}}|+|C^{\mathcal{I}}|-|I^{\mathcal{I}}|} \cdot \max\left(\frac{|I^{\mathcal{I}}|}{|D^{\mathcal{I}}|}, \frac{|I^{\mathcal{I}}|}{|C^{\mathcal{I}}|}\right) = s(D, C)$$

note that $I$ remains the same because of the commutativity of intersection.

### 5.2 Complexity Issues

In order to assess the complexity of $s$, the three different cases of applicability of the measure are discussed separately. They all depend on the complexity of the instance checking inference for the adopted DL language, hereafter indicated with $C(\mathsf{IC})$.

**Similarity between concepts:** $s$ is a numerical measure, all calculus in $s$ need of constant complexity; it holds that:

$$C(s) = 3 \cdot C(\mathsf{IC})$$

because the instance check is repeated three times: for the concept descriptions $C$, $D$ and $I$.

**Similarity between an individual and a concept:** in this case, besides of the instance checking operations required by the previous case, the MSC approximation of the considered individual is to be computed. Thus, denoted with $C(\mathsf{MSC}^*)$ the complexity of the MSC approximation, it holds that:

$$C(s) = C(\mathsf{MSC}^*) + 3 \cdot C(\mathsf{IC})$$

**Similarity between individuals:** this case is analogous to the previous one, the only difference is that now two MSC* approximations are to be computed for the arguments. So the complexity in this case is:

$$C(s) = 2 \cdot C(\mathsf{MSC}^*) + 3 \cdot C(\mathsf{IC})$$

As clearly shown by these formulæ, the measure complexity is sensible to the choice of the reference DL. For instance, for the $\mathcal{ALC}$ logic, $C(\mathsf{IC})$ is PSPACE (see [1], Ch. 3). For the cases involving individuals it suffices to recall that also the computation of the MSC approximations depends on instance checking besides of the specific algorithm [23].

## 6 Conclusions and Future Work

We have introduced a new similarity measure $s$ which is primarily meant for computing a similarity value between concept descriptions but, as previously shown, it could be also employed for assessing the similarity between individuals and between a concept description and an individual.

As previously suggested, such a measure could be applied to many tasks, namely clustering and retrieval on DL knowledge bases.

This measure could be improved in the case of non overlapping concepts. In particular, we will try to assess similarity in different cases employing a notion of *distance* between concepts.

This is strictly related to the weakness of the presented semantic similarity measure in cases involving individuals. In particular, we are addressing our research on the tuning of a semantic operator for the generalization of the approximated MSC obtained. In this way, we may overcome the actual problem, because the concept would be less specific then MSC and so it would instantiate more individuals that the selected one for calculating the similarity value. With this fitting we would keep a totally semantic similarity measure, but really applicable in every context concerning DLs representations.